\documentclass[runningheads]{llncs}

\usepackage{eccv}

\usepackage{eccvabbrv}

\usepackage{graphicx}
\usepackage{booktabs}

\usepackage[accsupp]{axessibility}  
\usepackage{algorithm}
\usepackage{algorithmicx}
\usepackage{algpseudocode}
\usepackage{bbm}
\usepackage{multirow}
\usepackage{multicol}
\renewcommand{\thefootnote}{\fnsymbol{footnote}}

\usepackage{hyperref}

\usepackage{orcidlink}

\begin{document}

\def\ourmodel{QASA}
\title{QASA: Quality-Guided K-Adaptive Slot Attention for Unsupervised Object-Centric Learning}

\titlerunning{QASA}

\author{
Tianran Ouyang\inst{1} \and
Xingping Dong\inst{1}\thanks{Corresponding author.} \and
Jing Zhang\inst{2} \and
Mang Ye\inst{1} \and
Kaihao Zhang\inst{2} \and
Bo Du\inst{1}
}

\authorrunning{T. Ouyang et al.}

\institute{
School of Computer Science, National Engineering Research Center for Multimedia Software, Institute of Artificial Intelligence, Hubei Key Laboratory of Multimedia and Network Communication Engineering, Wuhan University, Wuhan, China\\
\and
Australian National University, Canberra, Australia\\
\email{\{ouyangtianran,xingpingdong,yemang,dubo\}@whu.edu.cn}
\email{jing.zhang@anu.edu.au, super.khzhang@gmail.com}
}

\maketitle
\setcounter{footnote}{0}
\renewcommand{\thefootnote}{\arabic{footnote}}

\begin{abstract}
Slot Attention, an approach that binds different objects in a scene to a set of "slots", has become a leading method in unsupervised object-centric learning.
Most methods assume a fixed slot count $K$, and to better accommodate the dynamic nature of object cardinality, a few works have explored $K$-adaptive variants.
However, existing $K$-adaptive methods still suffer from two limitations. First, they do not explicitly constrain slot-binding quality, so low-quality slots lead to ambiguous feature attribution. Second, adding a slot-count penalty to the reconstruction objective creates conflicting optimization goals between reducing the number of active slots and maintaining reconstruction fidelity.
As a result, they still lag significantly behind strong $K$-fixed baselines.
To address these challenges, we propose \textbf{Q}uality-Guided $K$-\textbf{A}daptive \textbf{S}lot \textbf{A}ttention (\textbf{\ourmodel}).
First, we decouple slot selection from reconstruction, eliminating the mutual constraints between the two objectives.
Then, we propose an unsupervised Slot-Quality metric to assess per-slot quality, providing a principled signal for fine-grained slot--object binding.
Based on this metric, we design a Quality-Guided Slot Selection scheme that dynamically selects a subset of high-quality slots and feeds them into our newly designed gated decoder for reconstruction during training.
At inference, token-wise competition on slot attention yields a $K$-adaptive outcome.
We conduct experiments on both object discovery and object property prediction.
Results show that \ourmodel\ substantially outperforms existing $K$-adaptive methods on both real and synthetic datasets.
Moreover, on real-world datasets, \ourmodel\ surpasses $K$-fixed methods.
The code is available at \url{https://github.com/ouyangtianran/QASA-tianran}.

\end{abstract}

    \begin{figure}[thb!]
        \flushleft
        \includegraphics[width = 0.99 \textwidth]{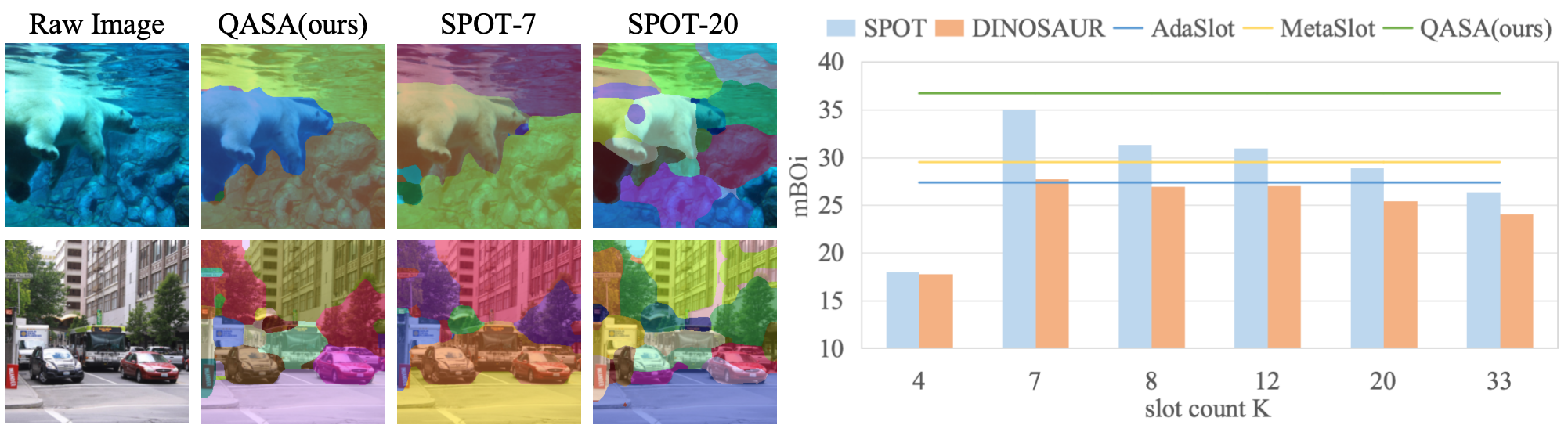}
        \caption{\textbf{mBOi vs.\ slot count $K$ on COCO~\cite{lin2014microsoft}.} $K$-fixed methods, such as SPOT~\cite{kakogeorgiou2024spot} and DINOSAUR~\cite{seitzer2023bridging}, show large performance fluctuations as $K$ varies.
        By contrast, $K$-adaptive methods do not rely on a carefully tuned $K$.
        Our method substantially outperforms existing $K$-adaptive baselines, including MetaSlot~\cite{liu2025metaslot} and AdaSlot~\cite{fan2024adaptive}, and even surpasses strong $K$-fixed baselines.}
        \label{fig:K-change}
    \end{figure}

\section{Introduction}
\label{sec:intro}

Object-centric learning (OCL) seeks to decompose scenes into sets of object representations that align with the causal structure of the physical world.
Because scenes are compositional, such representations can improve robustness to distribution shifts~\cite{dittadi2021generalization} and support downstream tasks such as reasoning and control~\cite{assouel2022object,mambelli2022compositional,biza2022binding,zadaianchuk2021self}.
Motivated by the abundance of unlabeled images, we focus on \textit{unsupervised object-centric learning}, where models learn object-centric representations directly from visual data.
Most unsupervised OCL methods optimize reconstruction objectives with set-structured bottlenecks that map an image to a small set of slots, which are then decoded back to the input space~\cite{locatello2020object}.
Within this line, Slot Attention has emerged as a mainstream approach~\cite{seitzer2023bridging,fan2024adaptive,tian2025pay,zhao2025multi,singh2025glass}.

Most current Slot Attention approaches fix the slot count $K$ a priori. However, the number of objects varies dynamically across different images, making object--slot mismatches inevitable. As shown in Fig.~\ref{fig:K-change}, for $K$-fixed methods, model performance is highly sensitive to the chosen value of $K$. The contradiction between a fixed $K$ and dynamically varying object counts hinders the further development of such techniques.
A pioneering $K$-adaptive approach, AdaSlot~\cite{fan2024adaptive}, trains a slot selector by adding a slot-count penalty to the training objective. However, the reconstruction objective tends to use more slots, which conflicts with the slot-count penalty. Other lines of work perform slot selection via principal component analysis (CoSA~\cite{kori2024grounded}) or by maintaining a unified slot codebook (MetaSlot~\cite{liu2025metaslot}). However, all these methods directly or indirectly rely on the prominence of object features, and thus tend to select only the most prominent objects for binding. This reliance on prominence leads to ambiguous slot--object attribution, where slots fail to bind precisely to specific objects and instead capture only coarse trends. Empirically, existing $K$-adaptive methods lag far behind $K$-fixed baselines, particularly in real-world scenarios.
These issues limit the practical adoption of $K$-adaptive Slot Attention.

To address these challenges, we propose \textbf{Q}uality-Guided $K$-\textbf{A}daptive \textbf{S}lot \textbf{A}ttention (\textbf{\ourmodel}). Unlike prior approaches that train a small auxiliary network to select slots during reconstruction, \ourmodel~decouples slot selection from reconstruction. 
To select high-quality slots under unsupervised conditions, we introduce a novel Slot-Quality metric. Inspired by the competitive token assignment in Slot Attention~\cite{locatello2020object, seitzer2023bridging}, we define slot's quality as the proportion of its attention mass within its winning region. Ideally, a high-quality slot wins entirely on its own tokens and assigns negligible mass elsewhere. 
Building on this metric, we propose a Quality-Guided Slot Selection scheme that, for each image, iteratively adds the highest-quality slots to a qualified set until a coverage threshold is met. We feed the slots into gated decoders, using gating to suppress the contributions of unselected slots, and compute the reconstruction loss.
For the object discovery task, we evaluate \ourmodel~on real-world datasets (COCO~\cite{lin2014microsoft} and PASCAL VOC~\cite{everingham2011pascal}) and synthetic datasets (MOVi-C and MOVi-E~\cite{greff2022kubric}). \ourmodel~substantially outperforms existing K-adaptive methods. Furthermore, it achieves state-of-the-art performance on real-world data relative to K-fixed methods. For the object property prediction task, \ourmodel~surpasses all baselines.
Our key contributions are as follows:
\begin{itemize}
\item We introduce a novel unsupervised \textbf{Slot-Quality} metric, defined as the proportion of its attention mass within its winning region, providing an effective signal to guide precise slot--object binding.
\item We propose a \textbf{Quality-Guided Slot Selection} module that jointly leverages quality, coverage, and novelty to guide the model to focus only on the most matching slot subset for each image.
\item We design \textbf{gated Transformer} and \textbf{gated MLP} decoders. Under both types of decoders, our method surpasses existing approaches, demonstrating the high compatibility of our proposed quality-guided mechanism.
\item Extensive experiments across object discovery and property prediction tasks demonstrate the broad superiority of \ourmodel. Specifically, it outperforms existing $K$-adaptive baselines by an average of \textbf{8.4\%} on mBOi for object discovery and achieves best results in the object property prediction task.
\end{itemize}

\section{Related work}
\label{sec:related}

\textbf{Object-Centric Learning.}
Object-centric learning (OCL) represents scenes as compositions of objects. Early approaches fall into two families: spatial-attention models that infer object bounding boxes~\cite{eslami2016attend,kosiorek2018sequential,crawford2019spatially} and scene-mixture models trained with variational inference~\cite{burgess2019monet,greff2019multi,engelcke2019genesis}.
Slot Attention~\cite{locatello2020object} crystallizes a set-structured bottleneck by iteratively assigning features to a set of slots and has become a mainstream paradigm in recent years.
Subsequent work has extended Slot Attention along four main directions:
(i) \emph{Reconstruction target}: As self-supervised pretrained models have become strong foundations for downstream tasks~\cite{caron2021dino,oquab2024dinov2,dong2022cfe,ouyang2025PL-CS}, DINOSAUR~\cite{seitzer2023bridging} replaces pixel reconstruction with pretrained feature reconstruction to better scale to real images.
(ii) \emph{decoders}: SLATE~\cite{singh2022illiterate} conditions an autoregressive/quantized decoder on slots to enhance detail and compositionality, and diffusion-based decoders further improve fidelity (e.g., LSD~\cite{jiang2023object} and SlotDiffusion~\cite{wu2023slotdiffusion});
(iii) \emph{training strategy}: SPOT~\cite{kakogeorgiou2024spot} leverages self-training with a patch-order permutation to strengthen slot pipelines;
(iv) \emph{video}: SAVi/SAVi++~\cite{kipf2022conditional,elsayed2022savi++} extend slot attention to videos, while VideoSAUR~\cite{zadaianchuk2023object} and SlotContrast~\cite{manasyan2025slotcontrast} further incorporate self-supervision.
However, these advances still assume a pre-specified slot count $K$. When $K$ mismatches scene complexity, under- and over-segmentation and wasted capacity occur, hampering learning and transfer~\cite{fan2024adaptive}.
We therefore treat $K$ as \emph{per-image adaptive}, retaining an image-specific set of slots.

\textbf{$K$-Adaptive Slot Attention.}
$K$-Adaptive Slot Attention aims to match model capacity to scene complexity instead of fixing $K$ globally.
Some $K$-fixed methods, such as SOLV~\cite{aydemir2023solv} and DIAS~\cite{zhao2025dias}, adjust the effective slot count by merging similar slots during inference, but cannot handle scenes with more objects than the preset $K$.
In contrast, $K$-adaptive methods use a large slot budget or candidate pool, providing stronger adaptability.
A representative $K$-adaptive method is AdaSlot~\cite{fan2024adaptive}, which starts from a large $K_{\max}$ and learns a selector with expected-$K$ regularization, together with a masked decoder.
Some methods expand the candidate slots by introducing external slot repositories.
CoSA~\cite{kori2024grounded} selects slots from a grounded slot dictionary via principal component analysis.
MetaSlot~\cite{liu2025metaslot} goes further by learning a slot codebook to enlarge the candidate set.
Outside the Slot Attention family, GENESIS-V2 employs a stochastic stick-breaking prior to infer a variable number of components~\cite{engelcke2021genesis}.
While $K$-adaptive designs have improved flexibility, they often favor prominent objects over fine structures and still trail $K$-fixed methods.
Our slot-quality criterion retains high-quality fine-grained slots, substantially narrowing this gap and surpassing $K$-fixed methods on complex scenes.

\begin{figure*}[thb!]
	\centering
	\includegraphics[width = 0.95 \textwidth]{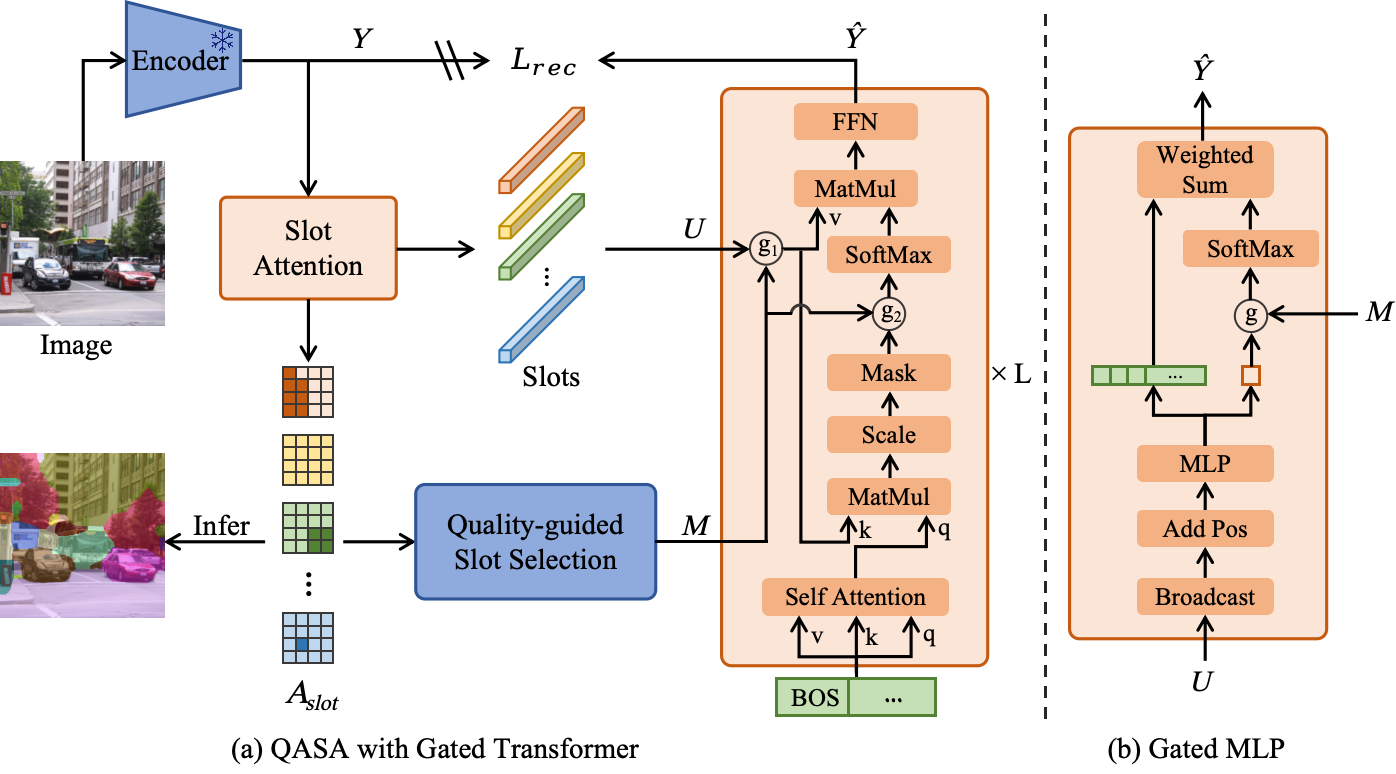}\\
	\caption{\textbf{Framework of Quality-Guided $K$-Adaptive Slot Attention (\ourmodel).}The overall training pipeline follows the canonical Slot Attention architecture: an image is encoded by the encoder, fed into the Slot Attention module, and then decoded; the reconstruction loss is computed against embeddings produced by a \emph{frozen} encoder. The key difference is that we compute a slot-quality metric from the Slot Attention maps $A^{\text{slot}}$ and use it to select a subset of slots, allowing only the selected slots to participate in decoding. At inference, no slot selection is applied. All slots compete via $A^{\text{slot}}$ for each token, yielding a $K$-adaptive result.
    }
	\label{fig:qasa}
\end{figure*}

\section{Method}
\label{sec:method}

\subsection{Overview}
\label{subsec:overview}

Slot Attention typically comprises three parts: \emph{encoder}, \emph{slot-attention} and \emph{decoder}. First, an encoder extracts $N$ patch-wise features $Y \in \mathbb{R}^{N \times d_y}$ from an input image $X$. Then, a slot-attention module groups these features into $K$ latent vectors $U = (u_1,\ldots,u_K) \in \mathbb{R}^{K \times d_u}$, referred to as \emph{slots}, each intended to represent one object in the image. Finally, a decoder reconstructs a target signal, either the original image $X$ or the features $Y$, from these slots.
Slot Attention is commonly evaluated using attention masks $A$ that indicate the association of each image patch with a specific slot. These masks can be obtained either from the slot-attention module or from the decoder, denoted $A^{\text{slot}}$ and $A^{\text{dec}}$, respectively. In this paper, we consistently use masks $A^{\text{slot}}$ for evaluation.

Prior $K$-adaptive methods mitigate the fixed-$K$ limitation through slot masking, selection, or pruning.
AdaSlot~\cite{fan2024adaptive} learns a selector with an expected-$K$ regularizer, but reconstruction naturally favors activating more slots, creating a conflict between reconstruction fidelity and slot-count reduction.
CoSA~\cite{kori2024grounded} decouples slot selection using a PCA-based criterion, while MetaSlot~\cite{liu2025metaslot} maintains a global codebook to quantize and prune redundant slots.
However, these criteria mainly aim to reduce slot usage or retain prominent components, without explicitly specifying what properties the retained slots should satisfy.
As a result, selected slots may still capture broad or diffuse regions rather than fine-grained object-specific structures with attention concentrated on the target object.
We instead introduce an unsupervised \emph{Slot-Quality} metric that evaluates whether a slot binds fine-grained target features within its winning region, providing an explicit criterion for selecting high-quality slots.
Building on this metric, we introduce \textbf{Q}uality-Guided $K$-\textbf{A}daptive \textbf{S}lot \textbf{A}ttention (\textbf{\ourmodel}), which selects slots through a metric-based procedure, thereby avoiding the contradiction introduced by slot-count regularization.
As shown in Fig.~\ref{fig:qasa}(a), \ourmodel~consists of four parts: \emph{encoder}, \emph{slot-attention}, \emph{slot selection} and \emph{gated decoder}.

During training, we set a maximum slot count $K_{\max}$, and we select an appropriate set of slots according to $A^{\text{slot}}$ (detailed in \S~\ref{subsec:q-select}) as $M = F_{\text{select}}\!\left(A^{\text{slot}}\right)$.
Then we feed the slots together with the selection mask into our proposed gated decoder (detailed in \S.~\ref{subsec:dec}), obtaining $\hat{Y} = Dec_{\text{gated}}(U,M)$.
Following DINOSAUR~\cite{seitzer2023bridging} and SPOT~\cite{kakogeorgiou2024spot}, we use DINO~\cite{caron2021emerging} as the image encoder and reconstruct its features $\hat{Y} \in \mathbb{R}^{N \times d_y}$ as the training target. The model is trained with the reconstruction loss: $\mathcal{L}_{\mathrm{rec}} = \frac{1}{N \cdot d_y}\lVert Y - \hat{Y} \rVert_{2}^{2}$.

During inference, no slot selection is applied and \emph{all} slots compete, based on $A^{\text{slot}}$, for each token to produce a $K$-adaptive result.
Concretely, for each patch $y_t \in Y$ we take a hard winner over the slot-attention map $A^{\text{slot}}$ via $w_t = argmax_{i \in \{1,\dots,K_{\max}\}} \; A^{\text{slot}}_{t,i}$,
and assign patch $y_t$ to the slot with the strongest association, i.e., $y_t$ is assigned to slot $w_t$. 
This induces a hard partition of tokens, $\mathcal{S}_i = \{\, t \mid w_t = i \,\}$, and we treat the union of patches indexed by $\mathcal{S}_i$ as the region assigned to slot $i$. Since the slot-attention masks are defined on the encoder's token grid (low resolution), we upsample them to the original image resolution before evaluating against ground-truth masks.

\subsection{Quality-Guided Slot Selection}
\label{subsec:q-select}

To perform training-free slot selection, the central challenge is to evaluate the quality of the slots in an unsupervised setting. The attention map of the slot-attention module $A^{\text{slot}}$ provides an excellent signal. Ideally, the attention of a high-quality slot $A^{\text{slot}}_i$ would be $1$ over its object region and $0$ elsewhere, but in the unsupervised setting the object regions are unknown. Inspired by Slot Attention's competitive assignment mechanism, in which attention maps induce competition among slots for responsibility over all tokens, we propose a competition-based slot-quality metric.

Let $A \in [0,1]^{N \times K_{\max}}$ denote the slot-attention probabilities for all tokens (patches) and slots, normalized over slots (i.e., $\sum_{i=1}^{K_{\max}} A_{t,i} = 1$ for each token $t$). For each token $t$, define its winner slot:
\begin{equation}
\label{eq:winner}
w_t \;=\; argmax_{i \in \{1,\ldots,K_{\max}\}} A_{t,i}.
\end{equation}
For slot $i$, define the winner mass and total mass:
\begin{equation}
\label{eq:W}
W_i^{\text{win}} \;=\; \sum_{t:\, w_t = i} A_{t,i},
\qquad
W_i \;=\; \sum_{t=1}^{N} A_{t,i}.
\end{equation}
The slot-quality score is
\begin{equation}
\label{eq:Q}
Q_i \;=\; \frac{W_i^{\text{win}}}{W_i + \epsilon} \;\in\; [0,1],
\end{equation}
which measures the purity of $slot_i$'s mass within the region where it wins.

Based on \emph{slot-quality}, we propose a Quality-Guided Slot Selection scheme that also considers \emph{coverage}, as detailed in Algorithm~\ref{alg:q-select}.
We set a \emph{per-token coverage-quality threshold} $\tau \in (0,1]$ and a \emph{coverage-rate threshold} $\rho \in (0,1]$. A token $t$ is marked as \emph{covered} if the cumulative attention mass from the selected slots $S \subseteq \{1,\ldots,K_{\max}\}$ exceeds $\tau$:
\begin{equation}
\label{eq:cov_token_simple}
\mathrm{Covered}_S(t) \;=\; \mathbbm{1}\!\left[\,\sum_{i \in S} A_{t,i} \;\ge\; \tau\,\right].
\end{equation}
The overall coverage rate is then
\begin{equation}
\label{eq:cov_rate_simple}
\mathrm{Coverage}(S) \;=\; \frac{1}{N}\sum_{t=1}^{N} \mathrm{Covered}_S(t).
\end{equation}

We greedily add slots in descending order of quality $Q_i$~\ref{eq:Q} until $\mathrm{Coverage}(S) \ge \rho$. To avoid repeatedly selecting slots that mostly explain already-covered regions, we enforce a \emph{novelty constraint}: a candidate slot $i$ is skipped if the fraction of its mass on already-covered tokens exceeds a threshold. Writing $C_S=\{\,t \mid \mathrm{Covered}_S(t)=1\,\}$ for the current covered set, the novelty of slot $i$ w.r.t.\ $S$ is
\begin{equation}
\label{eq:novelty}
\mathrm{novelty}(i \mid S) \;=\; 1 \;-\; \frac{\sum_{t \in C_S} A_{t,i}}{\sum_{t=1}^{n} A_{t,i} \;+\; \epsilon}\,.
\end{equation}
We accept $i$ only if $\mathrm{novelty}(i \mid S) \ge \mu$ with $\mu \in [0,1)$.
The resulting mask $M$ is used during training; no mask is applied at inference.

\begin{algorithm}[t]
\caption{Quality-Guided Slot Selection}
\label{alg:q-select}
\begin{algorithmic}[1]
\Require Slot-attention probabilities $A \in [0,1]^{N \times K_{\max}}$, thresholds $\tau,\rho,\mu$.
\Ensure Selection mask $M \in \{0,1\}^{K_{\max}}$.
\State \textbf{Compute qualities:} obtain $Q$ using Eq.~\ref{eq:Q} (with $w_t$ from Eq.~\ref{eq:winner} and $W_i^{\text{win}}, W_i$ from Eq.~\ref{eq:W}).
\State $S \leftarrow \varnothing$, \quad $M \leftarrow \mathbf{0}_{K_{\max}}$.
\State $\pi \leftarrow \operatorname*{argsort}_{i \in \{1,\dots,K_{\max}\}}(-Q_i)$ \Comment{indices in descending $Q$}
\For{$j=1$ to $K_{\max}$}
  \State $i \leftarrow \pi_j$.
  \If{$\mathrm{novelty}(i \mid S)$ from Eq.~\ref{eq:novelty} $< \mu$} \textbf{continue} \EndIf
  \State $S \leftarrow S \cup \{i\}$.
  \If{$\mathrm{Coverage}(S)$ from Eq.~\ref{eq:cov_rate_simple} (with $\mathrm{Covered}_S$ from Eq.~\ref{eq:cov_token_simple}) $\ge \rho$}
     \State \textbf{break}
  \EndIf
\EndFor
\State Set $M_i \leftarrow 1$ for all $i \in S$; \Return $M$.
\end{algorithmic}
\end{algorithm}

\subsection{Gated Decoder}
\label{subsec:dec}

After completing slot selection, we need to remove the influence of unselected slots during decoding. A straightforward approach is to dynamically gate all slots according to the mask, suppressing the contribution of unselected slots during decoding.
For a fair comparison, we simply augment the MLP~\cite{seitzer2023bridging} and Transformer~\cite{kakogeorgiou2024spot} decoders used in prior $K$-fixed methods with our gating mechanism, without introducing any additional network components.

\textbf{Gated Transformer Decoder.}
We adapt the standard Transformer decoder~\cite{kakogeorgiou2024spot} by introducing a dual gating mechanism into its cross-attention layers, as illustrated in Fig.~\ref{fig:qasa}(a). This mechanism leverages the selection mask $M \in \{0,1\}^{K_{\max}}$ from our quality-guided selection process (Algorithm~\ref{alg:q-select}) to create two distinct gates for fine-grained control over slot participation.

Given the selection mask $M$, we define a K/V gate $g_1 \in \{\varepsilon_1, 1\}^K_{\max}$ and a logits gate $g_2 \in \{\varepsilon_2, 1\}^K_{\max}$, where $0 < \varepsilon_1 , \varepsilon_2 < 1$. For each slot $i \in \{1, \dots, K_{\max}\}$, the gates are defined as
\begin{equation}
(g_1)_i = M_i + (1-M_i)\varepsilon_1, \quad (g_2)_i = M_i + (1-M_i)\varepsilon_2.
\end{equation}
Within each cross-attention block, the attention mechanism is modified as follows. First, the key ($K$) and value ($V$) vectors from the slots are multiplicatively modulated by the gate $g_1$ to physically suppress the information flow from unselected slots. Second, an additive bias based on the gate $g_2$ is applied to the attention logits before the softmax operation. This penalizes the scores of unselected slots, ensuring they do not win the attention competition. The full gated attention output is computed as:
\begin{equation}
\label{eq:gated_attn}
\text{Output} = \text{softmax}\left(\frac{Q(K \odot g_1)^\top}{\sqrt{d_k}} + \log(g_2)\right) (V \odot g_1),
\end{equation}
where $\odot$ denotes the element-wise product broadcasted across the feature dimension. This mechanism ensures unselected slots are robustly pruned from both the information pathway (via $g_1$) and the competitive attention allocation (via $g_2$).

\textbf{Gated MLP Decoder.}
The MLP decoder predicts per-slot reconstructions $\{\hat{y}_{i,t}\in\mathbb{R}^{d}\}_{i=1}^{K_{\max}}$ for tokens $t=1,\ldots,N$ together with mixture logits $\ell\in\mathbb{R}^{K_{\max}\times N}$ over the slot dimension. Given a binary selection mask $M\in\{0,1\}^{K_{\max}}$ (active set $S=\{i\mid M_i=1\}$), we apply hard gating at the logit level, as shown in Fig.~\ref{fig:qasa}(b), and then normalize only over active slots:
\begin{equation}
\label{eq:masked-logits-mlp}
\begin{aligned}
&\ell'_{i,t} =
\begin{cases}
\ell_{i,t}, & M_i = 1,\\[2pt]
-\infty,    & M_i = 0,
\end{cases} 
& where ~~ i \in \{1,\dots,K_{\max}\},\quad t \in \{1,\dots,N\}.
\end{aligned}
\end{equation}
\begin{equation}
\label{eq:softmax-active-mlp}
\alpha_{i,t}
\;=\;
\frac{\exp(\ell'_{i,t})}{\sum_{j=1}^{K_{\max}}\exp(\ell'_{j,t})}
\;=\;
\frac{M_i\,\exp(\ell_{i,t})}{\sum_{j=1}^{K_{\max}} M_j\,\exp(\ell_{j,t})}\,,
\end{equation}
The final reconstruction aggregates per-slot predictions using the gated mixture:
\begin{equation}
\label{eq:recon-mlp}
\hat{y}_t \;=\; \sum_{i=1}^{K_{\max}} \alpha_{i,t}\,\hat{y}_{i,t}\,, \qquad  t \in \{1,\dots,N\}.
\end{equation}
The masks reported for evaluation are $\alpha\in[0,1]^{K_{\max}\times N}$ reshaped to the token grid.
In practice, we implement the $-\infty$ gate by filling inactive-slot logits with a large negative constant $-\mathcal{C}$ (with $\mathcal{C}\gg 0$) before the softmax. Equivalently,
$\ell'_{i,t}=\ell_{i,t}-(1-M_i)\,\mathcal{C}$.
We ensure $|S|\ge 1$.

\section{Experiments}
\label{sec:exp}

\subsection{Setup}

\begin{figure*}[thb!]
	\centering
	\includegraphics[width = 0.99 \textwidth]{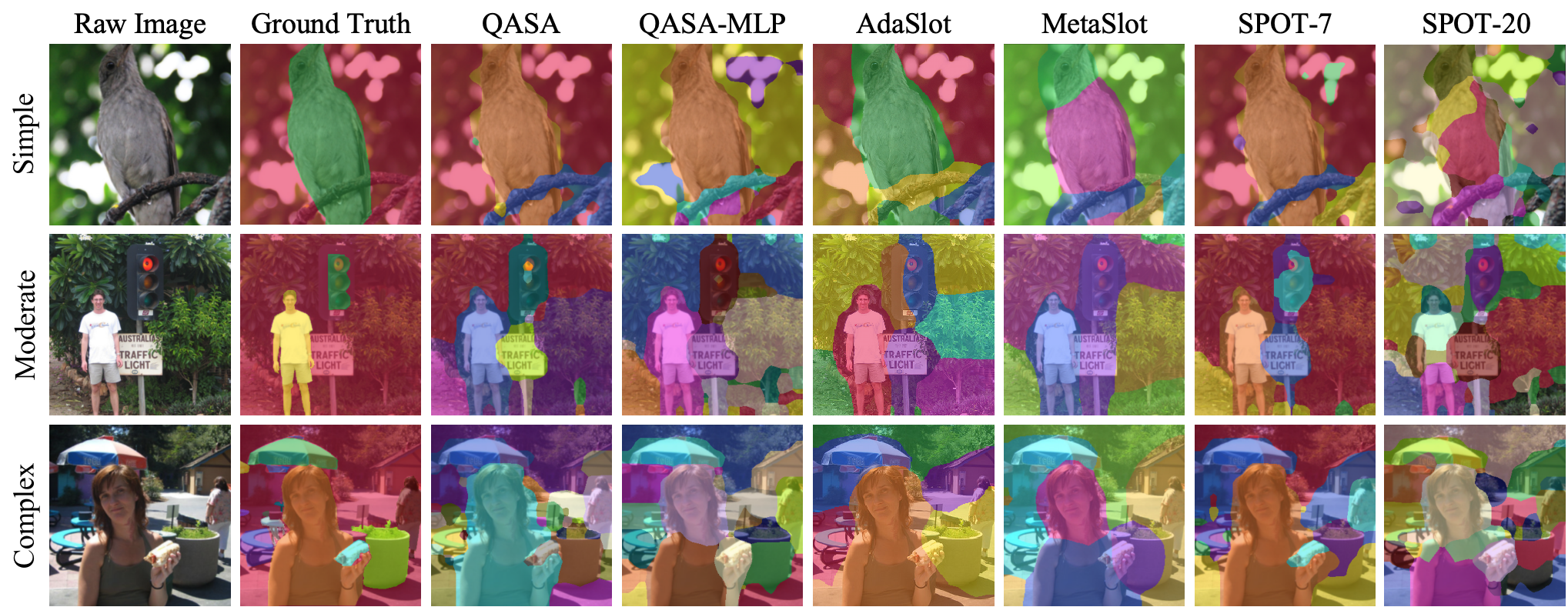}{
	\caption{\textbf{Visualizations on COCO.} The MetaSlot examples of AdaSlot~\cite{fan2024adaptive} and MetaSlot~\cite{liu2025metaslot} are generated using the released checkpoint. ''SPOT-7'' and ''SPOT-20'' denote SPOT~\cite{kakogeorgiou2024spot} with $K=7$ and $K=20$, respectively. The SPOT-7 examples are generated using the released checkpoint. The SPOT-20 examples are generated by training a model under the same settings as in the original paper.}
	\label{fig:example-coco}
    }
\end{figure*}

For object discovery, we evaluate on two widely used real-world datasets, \textbf{MS COCO 2017}~\cite{lin2014microsoft} and \textbf{PASCAL VOC 2012}~\cite{everingham2011pascal}, and two synthetic datasets, \textbf{MOVi-C} and \textbf{MOVi-E}~\cite{greff2022kubric}.
Our emphasis is on real-world scenes.
COCO contains diverse and cluttered images with many co-occurring objects and is particularly challenging, whereas VOC typically features one or a few salient objects and serves as a complementary regime.
For synthetic evaluation, MOVi provides rendered scenes with high-quality 3D-scanned assets.
MOVi-C has 3--10 objects per scene, and MOVi-E has 11--23.
For object property prediction, we follow the DINOSAUR and conduct experiments on COCO~\cite{seitzer2023bridging}.

\textbf{Metrics.} Following recent works~\cite{seitzer2023bridging,kakogeorgiou2024spot,fan2024adaptive,tian2025pay}, we report Mean Best Overlap (mBO) at the instance (mBOi) and class (mBOc) levels. For mBO, each ground-truth mask is paired with the single predicted mask that maximizes IoU, allowing many-to-one matches, and scores are then averaged per instance (mBOi) or per class (mBOc). We also report mean Intersection over Union (mIoU) under a stricter one-to-one protocol, where predicted and ground-truth masks are matched with the Hungarian algorithm before averaging. 
In line with previous works~\cite{tian2025pay,kakogeorgiou2024spot}, we do not report FG-ARI in our comparisons. FG-ARI ignores background regions and is insensitive to boundary quality, which can yield misleadingly optimistic segmentation scores~\cite{kakogeorgiou2024spot}.
For property prediction, we evaluate macro-averaged Top-1/Top-5 accuracy for classification and the coefficient of determination ($R^2$) for center coordinates ($R^2$-X, $R^2$-Y)~\cite{seitzer2023bridging}.

\textbf{Implementation Details.} 
Unless otherwise specified, all settings follow SPOT~\cite{kakogeorgiou2024spot}.
For all datasets, we set $\rho=0.8$ and $\mu=0.3$. For COCO and VOC, we set $\tau=0.5$. And for MOVi, we set $\tau=0.8$. 
For COCO and VOC, we set $K_{\max}$ to the maximum number of masks plus one (to include background), i.e., $K_{\max}=33$ and $K_{\max}=20$, respectively.\footnote{In the VOC training set, apart from a single sample with 38 masks, all others have at most 19 masks, so we ignore this outlier.} For MOVi-C and MOVi-E, we round the maximum mask count up to the nearest multiple of ten and use $K=20$ and $K=30$, respectively. 
We use $4\times$ RTX~4090 GPUs for COCO and $2\times$ RTX~4090 GPUs for the other datasets. With the Transformer decoder, the per-GPU batch size is 256. With the MLP decoder, the per-GPU batch size is 32. We train for 100 epochs on COCO and MOVi, and for 600 epochs on PASCAL~VOC. For training stability, we adopt a warm-up strategy in which all slots participate in reconstruction in the early epochs. With the Gated Transformer decoder, we set $warmup_{gate}=10$ on VOC and $20$ on MOVi-C/E. With the Gated MLP decoder, we set $warmup_{gate}=10$ on VOC, $10$ on MOVi-C, and $5$ on MOVi-E. On COCO, we disable this warm-up (i.e., $warmup_{gate}=0$). Detailed experimental settings are provided in the supplementary material.

\begin{table*}[t]
\centering
\caption{\textbf{Comparison of $K$-adaptive methods.} Results for AdaSlot and MetaSlot are taken from their original papers~\cite{fan2024adaptive, liu2025metaslot}. Dashes indicate results not reported.}
\label{tab:K-adaptive}
\resizebox{0.8\textwidth}{!}{
\begin{tabular}{l l l cc cc cc}
\toprule
\multicolumn{3}{c}{} & \multicolumn{2}{c}{COCO} & \multicolumn{2}{c}{VOC} & \multicolumn{1}{c}{MOVi-C} & \multicolumn{1}{c}{MOVi-E} \\
\cmidrule(lr){4-5}\cmidrule(lr){6-7}\cmidrule(lr){8-8}\cmidrule(lr){9-9}
Method & Dec. & $K$-type & mBOi & mIoU & mBOi & mIoU & mBOi & mBOi \\
\midrule
AdaSlot~\cite{fan2024adaptive}   & MLP    & adaptive & 27.4 & --   & --   & --   & 35.6 & 29.8 \\
MetaSlot~\cite{liu2025metaslot} & MLP    & adaptive & 29.5 & 27.9 & 43.9 & 42.1 & 35.0 & --   \\
\midrule
QASA                               & Trans. & adaptive & \textbf{36.7} & \textbf{35.0} & \textbf{49.7} & \textbf{47.9} & \textbf{46.9} & \textbf{39.1} \\
QASA$_{\text{MLP}}$                & MLP    & adaptive & 33.6 & 32.5 & 47.3 & 45.5 & 42.2 & 34.6 \\
\bottomrule
\end{tabular}
}
\end{table*}

\subsection{Object Discovery}

\paragraph{Comparison with $K$-adaptive methods.}

\begin{figure*}[thb!]
	\centering
	\includegraphics[width = 0.99 \textwidth]{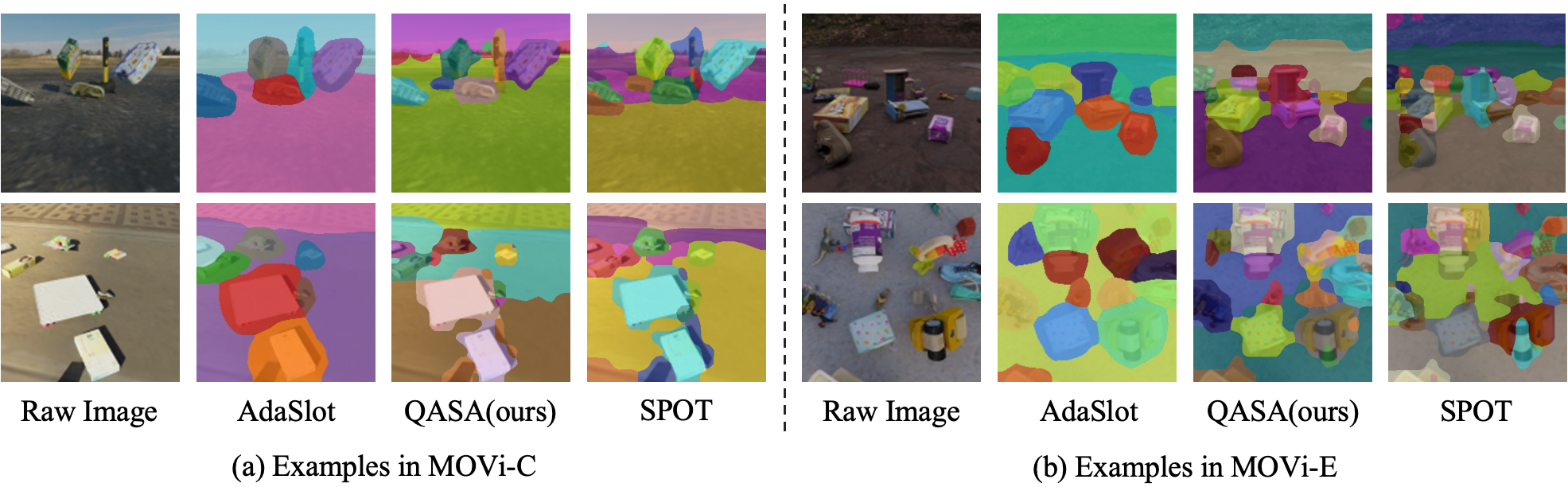}{\\
	\caption{\textbf{Visualizations on MOVi-C/E.} The AdaSlot examples are taken from its original paper~\cite{fan2024adaptive}. the SPOT examples are generated by training a model under the same settings as in the original paper.}
	\label{fig:example-movi}
    }
\end{figure*}

To the best of our knowledge, AdaSlot~\cite{fan2024adaptive}, CoSA~\cite{kori2024grounded}, and MetaSlot~\cite{liu2025metaslot} are the only three $K$-adaptive Slot Attention methods to date. 
Because CoSA lacks publicly available experimental artifacts for comparison, we report results against AdaSlot and MetaSlot only.
As shown in Tab.~\ref{tab:K-adaptive}, our method \ourmodel~outperforms AdaSlot and MetaSlot by a large margin on both real-world (COCO) and synthetic (MOVi) datasets, with an average improvement of \textbf{8.4\%} in mBOi. Even when using an MLP decoder to match their setup, \ourmodel$_{\text{MLP}}$ still surpasses them substantially, achieving a \textbf{4.7\%} gain in mBOi. 
We present visualizations stratified by scene complexity to intuitively demonstrate the advantages of \ourmodel~on COCO, as shown in Fig.~\ref{fig:example-coco}. It is clear that \ourmodel~outperforms AdaSlot~\cite{fan2024adaptive} and MetaSlot~\cite{liu2025metaslot} across all three levels (simple, moderate, and complex). \ourmodel~adapts to varying object counts and captures fine-grained object details, whereas AdaSlot largely partitions images into coarse blob-like regions. This issue is especially pronounced in complex scenes.
MetaSlot can capture certain fine details, but its performance degrades markedly in complex scenes. On the synthetic datasets~\cite{greff2022kubric}, our method likewise outperforms AdaSlot~\cite{fan2024adaptive}, as shown in Fig.~\ref{fig:example-movi}, capturing sharper object boundaries rather than merely coarse blob-like outlines.
The visualizations show that \ourmodel~learns slots that capture fine object details and bind more precisely to objects, while preserving the ability to adapt the number of objects per image.

\begin{table*}[th]
\centering
\caption{\textbf{Comparison with $K$-fixed methods on COCO~\cite{lin2014microsoft}, PASCAL~VOC~\cite{everingham2011pascal}, and MOVi-C/E~\cite{greff2022kubric}.} Results for SLATE, DINOSAUR, DINOSAUR$_{\text{MLP}}$, and SPOT are taken from SPOT~\cite{kakogeorgiou2024spot}. Results for the remaining methods are taken from their original papers~\cite{tian2025pay}. Dashes indicate results not reported.}
\label{tab:k-fixed}
\resizebox{0.99\textwidth}{!}{
\begin{tabular}{l l l cc cccccccc}
\toprule
\multicolumn{3}{c}{} & \multicolumn{3}{c}{COCO} & \multicolumn{3}{c}{VOC} & \multicolumn{2}{c}{MOVi-C}& \multicolumn{2}{c}{MOVi-E}\\
\cmidrule(lr){4-6}\cmidrule(lr){7-9} \cmidrule(lr){10-11}\cmidrule(lr){12-13}
Method & Dec. & $K$-type & mBOc & mBOi & mIoU & mBOc & mBOi & mIoU & mBOi & mIoU & mBOi & mIoU\\
\midrule
SLATE~\cite{singh2022illiterate}   & --  & --  & 33.6  & 29.1 & --   & 41.5 & 35.9 & -- & 39.4 & 37.8  & 30.2  & 28.6 \\
DINOSAUR~\cite{seitzer2023bridging}  & Trans. & fixed  & 39.7 & 31.6 & -- & 51.2 & 44.0 & -- & 42.4 & --  & -- & -- \\
DINOSAUR$_{\text{MLP}}$~\cite{seitzer2023bridging}  & MLP & fixed & 30.9 & 27.7 & --  & 40.9 & 39.5 & -- & 39.1 & -- & 35.5 & -- \\
SPOT~\cite{kakogeorgiou2024spot}  & Trans. & fixed  & 44.7 & 35.0 & 33.0 & 55.6 & 48.3 & -- & 47.3 & 46.7 & \textbf{40.1} & \textbf{39.3}\\
SPOT+FS+RC~\cite{tian2025pay}   & Trans. & fixed    & 45.3 & 35.7 & --  & 56.5 & 49.3  & --  & \textbf{49.0}  & \textbf{47.8}  & --  & --  \\
DINOSAUR+FS+RC~\cite{tian2025pay}     & MLP & fixed    & 40.1 & 31.2 & --  & 52.9 & 45.2 & -- & 46.0 & 44.8 & -- & -- \\
\midrule
QASA          & Trans. & adaptive & \textbf{45.5} & \textbf{36.7} & \textbf{35.0} & \textbf{57.9} & \textbf{49.7} & \textbf{47.9} & 46.9 & 46.1 & 39.1 & 37.8 \\
QASA$_{\text{MLP}}$      & MLP    & adaptive & 36.5 & 33.6  & 32.5 & 53.8 & 47.3 & 45.5 & 42.2 & 41.1 & 34.6 & 33.0 \\
\bottomrule
\end{tabular}
}
\end{table*}

To more directly illustrate the advantage of \ourmodel, we further evaluate representative $K$-fixed methods (SPOT~\cite{kakogeorgiou2024spot} and DINOSAUR~\cite{seitzer2023bridging}) on COCO under varying $K$ (Fig.~\ref{fig:K-change}).
As $K$ changes, their performance fluctuates markedly. This sensitivity necessitates exhaustive per-dataset $K$ sweeps, which are time- and compute-intensive, yet these sweeps still do not resolve the structural mismatch between a fixed slot budget and image-dependent object counts.
Existing $K$-adaptive methods (AdaSlot~\cite{fan2024adaptive} and MetaSlot~\cite{liu2025metaslot}) can adaptively select the number of slots used per image, which saves substantial tuning effort and resources. However, they show a clear performance gap compared with $K$-fixed methods because they lack the slot-count prior.
By contrast, \ourmodel~not only adaptively selects appropriate slots for each image for reconstruction, but also surpasses $K$-fixed methods in performance, thereby addressing the shortcoming of $K$-adaptive methods.

\paragraph{Comparison with $K$-fixed methods.}
Tab.~\ref{tab:k-fixed} summarizes recent $K$-fixed methods, including SLATE~\cite{singh2022illiterate}, DINOSAUR~\cite{seitzer2023bridging}, SPOT~\cite{kakogeorgiou2024spot}, and +FS+RC~\cite{tian2025pay}.
On the real-world datasets COCO and VOC, \ourmodel\ achieves \emph{state-of-the-art} results across all metrics, despite lacking the strong inductive bias provided by a preset $K$.
On the synthetic datasets, our method does not outperform all $K$-fixed methods.
We attribute this to the more concentrated object-count distribution in synthetic data, where a carefully chosen fixed $K$ provides larger gains, and additional analysis is provided in the appendix.
We also note that the comparison with +FS+RC~\cite{tian2025pay} is not directly comparable, since it uses an extra foreground background indicator learned with additional training.
Overall, unlike AdaSlot~\cite{fan2024adaptive} and MetaSlot~\cite{liu2025metaslot}, whose strong performance is largely confined to toy or synthetic settings, \ourmodel\ surpasses $K$-fixed methods on real data and therefore offers clear practical benefits.

\subsection{Object Property Prediction}
\label{sec:prediction}

\begin{table*}[t]
\centering
\caption{\textbf{Object property prediction results.} We report classification accuracy and $R^2$ scores using an MLP predictor and a linear predictor.
DINOSAUR~\cite{seitzer2023bridging} and MetaSlot~\cite{liu2025metaslot} are trained with their official code.
SPOT~\cite{kakogeorgiou2024spot} and AdaSlot~\cite{fan2024adaptive} are evaluated using slots extracted from their official checkpoints.
All baselines use the same backbone and evaluation setup as our model.}
\label{tab:set-prediction}
\resizebox{0.99\textwidth}{!}{
\begin{tabular}{l l l cccc cccc}
\toprule
\multicolumn{3}{c}{} & \multicolumn{4}{c}{MLP predictor} & \multicolumn{4}{c}{Linear predictor} \\
\cmidrule(lr){4-7}\cmidrule(lr){8-11}
Method & Dec. & $K$-type & class-top1 & class-top5 & $R^2$-X & $R^2$-Y & class-top1 & class-top5 & $R^2$-X & $R^2$-Y \\
\midrule
DINOSAUR~\cite{seitzer2023bridging} & Trans. & fixed & 47.1 & 75.5 & 55.4 & 59.1 & 39.7 & 68.0 & 48.8 & 52.0 \\
DINOSAUR$_{\text{MLP}}$~\cite{seitzer2023bridging} & MLP & fixed & 46.9 & 75.4 & 62.9 & 63.3 & 39.1 & 67.5 & 59.2 & 59.9 \\
SPOT~\cite{kakogeorgiou2024spot} & Trans. & fixed & 45.9 & 72.7 & 51.9 & 56.4 & 45.2 & 72.1 & 49.2 & 53.4 \\
AdaSlot~\cite{fan2024adaptive} & MLP & adaptive & 49.3 & 79.0 & 72.7 & 69.8 & 33.9 & 66.3 & 58.6 & 53.3 \\
MetaSlot~\cite{liu2025metaslot} & MLP & adaptive & 40.9 & 70.0 & 55.0 & 55.3 & 37.9 & 66.7 & 52.0 & 53.2 \\
\midrule
QASA & Trans. & adaptive & \textbf{59.6} & \textbf{85.0} & 67.9 & 71.7 & \textbf{57.3} & \textbf{82.3} & \textbf{67.9} & \textbf{69.9} \\
QASA$_{\text{MLP}}$ & MLP & adaptive & 58.9 & 84.1 & \textbf{74.0} & \textbf{76.7} & 46.7 & 76.7 & 62.3 & 56.5 \\
\bottomrule
\end{tabular}
}
\end{table*}

To further demonstrate the downstream potential of the slots learned by our method~\cite{mamaghan2025exploring}, we conduct experiments on the object property prediction task.
We evaluate DINOSAUR~\cite{seitzer2023bridging}, SPOT~\cite{kakogeorgiou2024spot}, AdaSlot~\cite{fan2024adaptive}, and MetaSlot~\cite{liu2025metaslot} as baselines, training and testing all methods under the same protocol, with details provided in the appendix.
Following prior work~\cite{seitzer2023bridging}, we conduct experiments on COCO using both MLP and linear predictors.
As shown in Tab.~\ref{tab:set-prediction}, our method consistently outperforms all baselines under both predictors. 
With the stronger MLP predictor, \ourmodel\ improves top-1 classification by 10.3\% and increases the average $R^2$ by 4.1\%.
These results indicate that \ourmodel\ learns a more informative scene representation, where each slot encodes more accurate object information, which supports its strong potential for downstream applications.

\subsection{Ablation Study}

\begin{table}[th]
\centering
\begin{minipage}[t]{0.62\linewidth}
\centering
\caption{\textbf{Influence of key components in Slot Selection.} Coverage, Quality, and Novelty denote the corresponding mechanisms in the Quality-Guided Slot Selection. All results are evaluated on COCO~\cite{lin2014microsoft}.}
\label{tab:ablation}
\begin{tabular}{c c c c c c}
\toprule
Coverage & Quality & Novelty & mBOc & mBOi & mIoU \\
\midrule
$\times$         & $\times$         & $\times$         & 25.9 & 25.5 & 24.9 \\
$\checkmark$     & $\times$         & $\times$         & 34.5 & 25.3 & 22.7 \\
$\checkmark$     & $\checkmark$     & $\times$         & 45.0 & 35.0 & 32.9 \\
$\checkmark$     & $\checkmark$     & $\checkmark$     & \textbf{45.5} & \textbf{36.7} & \textbf{35.0} \\
\bottomrule
\end{tabular}
\end{minipage}\hfill
\begin{minipage}[t]{0.35\linewidth}
\centering
\caption{\textbf{Ablation on Gates in the Transformer decoder.} All results are evaluated on COCO~\cite{lin2014microsoft}.}
\label{tab:gating_ablation}
\begin{tabular}{c c c c c}
\toprule
$g_1$ & $g_2$ & mBOc & mBOi & mIoU \\
\midrule
$\times$         & $\times$         & 25.9 & 25.5 & 24.9 \\
$\times$         & $\checkmark$     & 26.1 & 25.5 & 24.9 \\
$\checkmark$     & $\times$         & 43.1 & 33.2 & 30.9 \\
$\checkmark$     & $\checkmark$     & \textbf{45.5} & \textbf{36.7} & \textbf{35.0} \\
\bottomrule
\end{tabular}
\end{minipage}
\end{table}

For an in-depth analysis of the contributions of each component, we conduct several ablation studies. 
All ablations use the Transformer decoder. Unless otherwise specified, all hyperparameters remain unchanged.

\textbf{Components of Quality-Guided Slot Selection.} Starting from a model without gating, we progressively enable the components of the quality-guided slot selection (\emph{coverage}, \emph{quality}, and \emph{novelty}), as shown in Tab.~\ref{tab:ablation}. Enabling \emph{coverage} alone chiefly lifts mBOc but slightly harms mBOi and mIoU because diffuse/low-quality slots are admitted to meet the coverage target. Adding \emph{quality} yields the largest jump overall by prioritizing high-quality slots, improving one-to-one alignment and boundary tightness. Incorporating \emph{novelty} further gives a gain on mBOc and clear gains on mBOi/mIoU by filtering overlapping slots.

\textbf{Gates in the Transformer decoder.} We further ablate the dual gating mechanism, as reported in Tab.~\ref{tab:gating_ablation}. Using only $g_2$ (logit gating) fails to enforce effective masking, likely because unattenuated logits can overwhelm the negative bias. Using only $g_1$ (key/value gating) already yields strong control, and combining $g_1$ with $g_2$ leads to additional gains. Concretely, $g_1$ scales the key and value vectors of unselected slots toward zero, which effectively shrinks their attention logits, while $g_2$ further adds a large negative bias to these logits, providing an extra margin that suppresses the influence of unselected slots.

\begin{figure*}[thb!]
	\centering
	\includegraphics[width = 0.99 \textwidth]{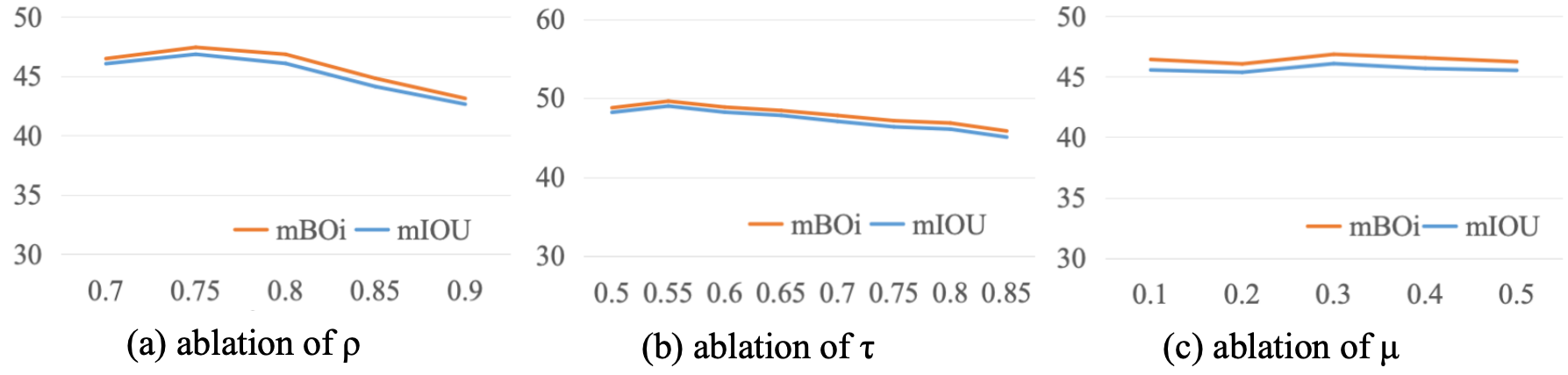}{\\
	\caption{\textbf{Influence of $\rho$, $\tau$ and $\mu$.} All results are evaluated on MOVi-C.}
	\label{fig:ablation}
    }
\end{figure*}

\textbf{Influence of $\rho$, $\tau$, and $\mu$.}
Fig.~\ref{fig:ablation} shows how mBOi and mIoU vary with the coverage-rate threshold $\rho$, the per-token coverage-quality threshold $\tau$, and the novelty threshold $\mu$.
The threshold $\rho$ determines whether the selected slots achieve sufficient spatial coverage and is therefore the basis of gating.
In practice, $\rho$ around 0.7 to 0.8 already yields reliable gating.
The threshold $\tau$ controls the token-level covered decision.
Once $\rho$ is fixed, performance is largely insensitive to $\tau$ over a wide range.
Adding novelty consistently improves performance, and the gains are stable across a wide range of $\mu$.
We do not tune these thresholds per dataset, since minor changes from fine tuning are not our focus, and the current results are sufficient to demonstrate the advantages of our model.

\begin{table*}[ht]
\centering
\caption{Effect of $K_{\max}$ on MOVi-C and MOVi-E~\cite{greff2022kubric}.}
\label{tab:Kmax-change}
\setlength{\tabcolsep}{6pt}
\begin{tabular}{c c c c c c c c}
\toprule
\multirow{3}{*}{MOVi-C} & $K_{\max}$ & 15 & 18 & 20 & 22 & 24 & 32 \\
& mBOi & 44.4 & 46.5 & 46.9 & 46.4 & 46.3 & 45.4 \\
& mIoU & 43.4 & 45.7 & 46.1 & 45.6 & 45.7 & 45.0 \\
\midrule
\midrule
\multirow{3}{*}{MOVi-E} & $K_{\max}$ & 24 & 28 & 30 & 32 & 34 & 64 \\
& mBOi & 38.3 & 39.1 & 39.1 & 39.3 & 39.5 & 38.7 \\
& mIoU & 36.8 & 37.8 & 37.8 & 38.0 & 38.3 & 38.3 \\
\bottomrule
\end{tabular}
\end{table*}

\textbf{Choice of $K_{\max}$.} Unlike $K$-fixed methods that require searching for an optimal $K$, $K$-adaptive methods only specify $K_{\max}$ and adaptively select effective slots per image. In practice, setting $K_{\max}$ to the maximum number of objects plus one (for background) usually works well. However, on the synthetic datasets MOVi-C and MOVi-E, we observe that moderately increasing $K_{\max}$ can yield better results, as shown in Tab.~\ref{tab:Kmax-change}. The gains from such slot redundancy are noticeably smaller on MOVi-E than on MOVi-C. We hypothesize this is because synthetic backgrounds are highly homogeneous, causing several slots to specialize to a few background modes. Increasing the number of redundant slots therefore helps. Since MOVi-E has more complex backgrounds than MOVi-C, this effect is reduced.
Importantly, even without such redundancy ($K_{\max}{=}15$ for MOVi-C and $K_{\max}{=}24$ for MOVi-E), \ourmodel~still performs strongly. The table also shows that \ourmodel~is highly robust to the choice of $K_{\max}$: performance remains stable over a wide range, and even when $K_{\max}$ is set far beyond the maximum object count ($K_{\max}{=}32$ for MOVi-C and $K_{\max}{=}64$ for MOVi-E), the degradation is minor, further supporting the effectiveness of the proposed $K$-adaptive mechanism.

\section{Conclusion}
\label{sec:conclusion}

We introduce a \emph{Slot-Quality} metric that drives slots toward fine, object-specific features. Building on this metric, we present \ourmodel, a $K$-adaptive Slot Attention method that selects the appropriate slots \emph{per image}, better handling uncertainty in object cardinality. We decouple slot selection from decoding, removing the conflict between slot-count regularization and reconstruction. The framework supports both Transformer and MLP decoders.
On object discovery, \ourmodel\ markedly outperforms existing $K$-adaptive methods, with up to an 8.4\% average improvement on mBOi.
Without hand-tuning an optimal $K$, \ourmodel\ achieves state-of-the-art performance on real-world datasets and outperforms $K$-fixed methods.
On object property prediction, \ourmodel\ substantially surpasses all baselines on both classification and position regression, demonstrating strong downstream potential.
With a simple architecture, \ourmodel\ dynamically determines the slot count from each scene and learns more informative and structured representations, which further highlights the promise of $K$-adaptive OCL.

\section*{Acknowledgments}
\begin{sloppypar}
This work was supported by the Fundamental and Interdisciplinary Disciplines Breakthrough Plan of the Ministry of Education of China (No.~JYB2025XDXM101), the National Natural Science Foundation of China (Grant No.~62471342), the Fundamental Research Funds for the Central Universities (No.~2042026kf0044), the New Cornerstone Science Foundation through the XPLORER PRIZE, and the Innovative Research Group Project of Hubei Province under Grant No.~2024AFA017.
\end{sloppypar}

%
%
\bibliographystyle{splncs04}
\bibliography{main}

\clearpage
\setcounter{page}{1}
\begin{center}
    \Large \textbf{Supplementary Material}
\end{center}
\appendix
\renewcommand{\thesubsection}{\Alph{section}.\arabic{subsection}}

\section{Implementation Details}
\label{app:details}

For a fair comparison and to demonstrate the effectiveness of our method, we align our hyperparameters and settings as closely as possible with the latest open-source baseline, SPOT~\cite{kakogeorgiou2024spot}. Details are provided below.

\textbf{Training.} We employ the Adam optimizer~\cite{diederik2015adam} with $\beta_1=0.9$, $\beta_2=0.999$, and no weight decay. The learning rate follows a linear warm-up from 0 to a peak value for 10000 iterations and then decreases via a cosine annealing schedule. For COCO and PASCAL, the peak learning rate is $4\times10^{-4}$ and the minimum learning is $4\times10^{-7}$. For MOVi, the peak learning rate is $2\times10^{-4}$ and the minimum learning is $4\times10^{-5}$.
We use $4\times$ RTX~4090 GPUs for COCO and $2\times$ RTX~4090 GPUs for the other datasets. With the Transformer decoder, the per-GPU batch size is 256. With the MLP decoder, the per-GPU batch size is 32. We train for 100 epochs on COCO and MOVi, and for 600 epochs on PASCAL~VOC. Dataset selection, preprocessing, and size configurations are identical to SPOT~\cite{kakogeorgiou2024spot} and are therefore omitted for brevity.

\textbf{Encoder.} We implement our QASA using a ViT-B/16 encoder~\cite{dosovitskiy2020image} initialized with DINO~\cite{caron2021emerging}, and we omit the final layer norm.

\textbf{Slot Attention.}
For all experiments, we use 3 Slot Attention iterations with slot dimension $d_u=256$ and a slot-attention MLP hidden dimension of 1024.
For COCO and VOC, we set $K_{\max}$ to the maximum number of masks plus one for background, yielding $K_{\max}=33$ and $K_{\max}=20$, respectively.
In the VOC training set, except for one sample with 38 masks, all others have at most 19 masks.
For MOVi-C and MOVi-E, we round the maximum mask count up to the nearest multiple of ten and use $K_{\max}=20$ and $K_{\max}=30$, respectively.

\textbf{Slot Selection.} We set $\rho=0.8$ and $\mu=0.3$ for all datasets. For COCO and VOC, we use $\tau=0.5$. For MOVi, we use $\tau=0.8$. To stabilize training, we adopt a warm-up strategy: during the first $warmup_{gate}$ epochs, no mask is applied and all slots participate in reconstruction. With the Transformer decoder, we set $warmup_{gate}=10$ on VOC and $20$ on MOVi-C/E. With the MLP decoder, we set $warmup_{gate}=10$ on VOC, $10$ on MOVi-C, and $5$ on MOVi-E. On COCO, we disable this warm-up (i.e., $warmup_{gate}=0$).

\textbf{Gated Transformer Decoder.} We use 4 Transformer blocks, each with 6 heads. Consistent with SPOT~\cite{kakogeorgiou2024spot}, we adopt its sequence-permutation strategy, using random permutations during training and all permutations at inference, but we do not use parallel decoding~\cite{tschannen2023image}. Given the selection mask $M$, we use a K/V gate $g_1 \in \{\varepsilon_1,1\}^{K_{\max}}$ and a logits gate $g_2 \in \{\varepsilon_2,1\}^{K_{\max}}$. We set $\varepsilon_1=10^{-6}$ and $\varepsilon_2=10^{-3}$ in all experiments.

\textbf{Gated MLP Decoder.}
For the MLP-based decoder, we adopt the spatial broadcast mechanism from prior work~\cite{seitzer2023bridging,watters2019spatial}.
Each slot is expanded into $N$ tokens aligned with image patches, and we add learnable positional embeddings to preserve spatial identity.
The resulting tokens are processed independently by a four-layer MLP with ReLU activations and hidden size 2048.
For each slot, the decoder predicts both a reconstruction and an alpha map, with per-token output dimension $d_y+1$, where $d_y$ corresponds to reconstruction channels and the extra channel corresponds to alpha.
The alpha map acts as an attention map that indicates the active regions of the slot, and the final reconstruction is obtained by aggregating per-slot reconstructions using the alpha maps as weights.
In this MLP decoder, the predicted alpha maps are also used as the slot-attention masks.
Given the selected-slot mask $M$, we apply hard gating at the alpha-logit level by filling logits of inactive slots with a large negative constant $C=-10^{9}$ before the softmax.

\section{Details on Object Property Prediction}
\label{app:property-prediction}

We investigate the utility of our learned representations for downstream applications by predicting object properties on the COCO dataset.

\textbf{Setup.} We follow the experimental design of DINOSAUR~\cite{seitzer2023bridging}. We extract slot representations using our unsupervised model (frozen) and use them to supervise a predictor for object category and center-of-mass coordinates ($x, y$).

\textbf{Baselines.}
Existing OCL works do not follow a unified protocol for this task.
Many papers do not evaluate on this task, or compare against only one baseline.
Backbone choices and evaluation metrics also vary across works.
To ensure fair comparison, we adopt a single shared setup and evaluate DINOSAUR~\cite{seitzer2023bridging}, SPOT~\cite{kakogeorgiou2024spot}, AdaSlot~\cite{fan2024adaptive}, and MetaSlot~\cite{liu2025metaslot} as baselines, training and testing all methods under the same protocol.
These baselines are representative, covering both $K$-fixed and $K$-adaptive approaches, including established and recent methods, and spanning both MLP and Transformer decoders.
Together, they provide a comprehensive assessment of property prediction performance.

\textbf{Architecture.}
We evaluate both MLP and linear predictors.
For the MLP predictor, we use a single-hidden-layer MLP with a hidden dimension of 256.
It operates per slot, applying shared weights independently to each slot representation in parallel to produce a vector of property predictions.

\textbf{Loss and Matching.} We employ cross-entropy loss for discrete classification (80 COCO categories) and Mean Squared Error (MSE) for continuous coordinate regression. The total loss is the sum of these terms. To align the set of predictions with the set of ground truth objects, we utilize the Hungarian algorithm to minimize the total matching loss.

\textbf{Data and Training Details.} The training set consists of 10,000 images randomly sampled from COCO, with an additional 1,000 images reserved for validation. 
We train the predictor for 15,000 steps using the Adam optimizer with a learning rate of $1 \times 10^{-3}$ and a batch size of 64. Evaluation is performed on the standard 5,000 validation images, excluding those containing no objects. We report the macro-averaged Top-1 and Top-5 accuracy for classification, and the coefficient of determination ($R^2$) for $x$ and $y$ coordinates. We evaluate the checkpoint with the lowest validation loss. For fair comparison, we do not apply any truncation to the ground truth.

\section{Further Analysis}
\label{app:analysis}

\begin{figure*}[thb!]
	\centering
	\includegraphics[width = 0.85 \textwidth]{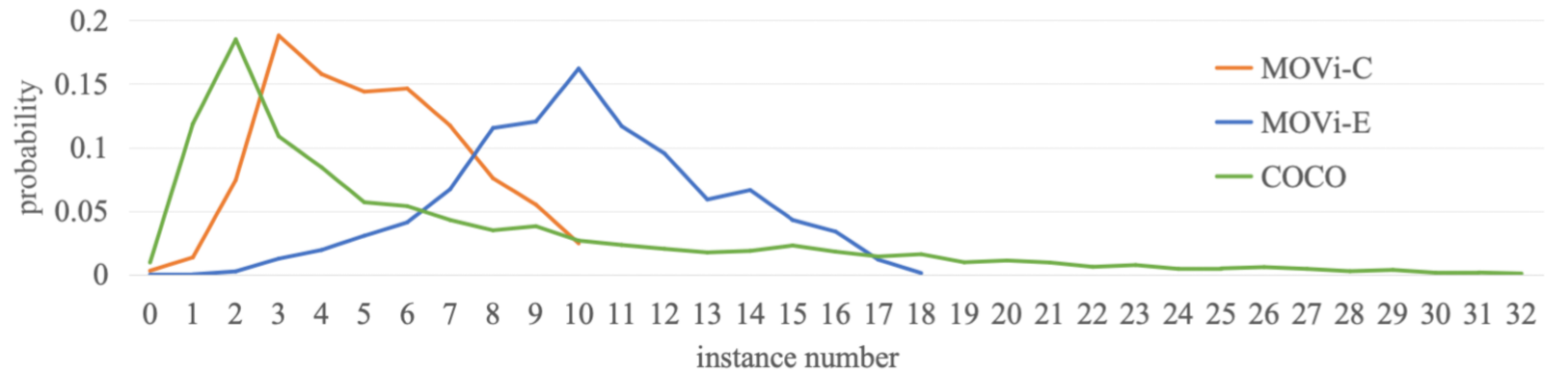}{\\
	\caption{\textbf{Object number distribution on COCO and MOVi-C/E.}}
	\label{fig:dataset-dis}
    }
\end{figure*}

\textbf{Dataset distribution.}
We analyze the object-count distribution of the validation sets on COCO and MOVi-C/E, as shown in Fig.~\ref{fig:dataset-dis}.
The synthetic datasets MOVi-C/E exhibit a more concentrated distribution with a narrower range of object counts.
This property provides a strong inductive bias for choosing a fixed $K$, which benefits $K$-fixed methods.
In contrast, COCO shows a more long-tailed distribution with a much wider range of object counts, which weakens the bias of a fixed $K$.
This dataset discrepancy helps explain why our method shows more pronounced gains on COCO.
It also suggests that our model has greater potential in real-world settings where object cardinality varies substantially.

\begin{figure}[h]
        \centering
        \includegraphics[width = 0.85\textwidth]{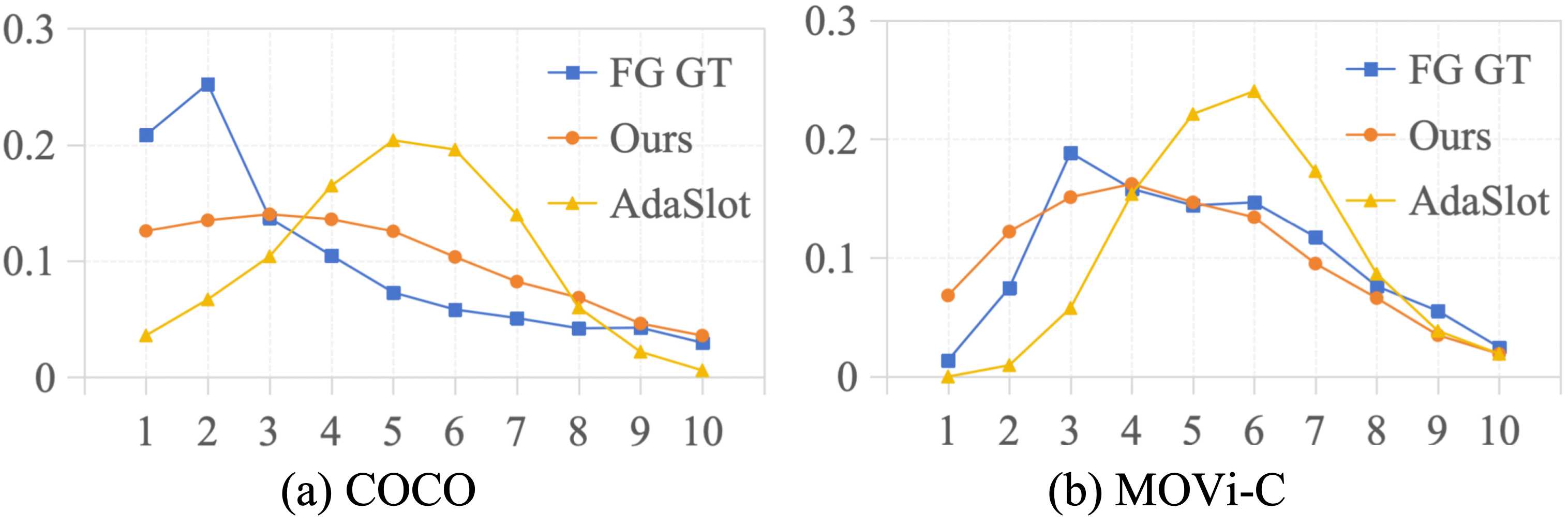}
        \caption{\textbf{Active slot count distribution.}}
        \label{fig:act-distri}
    \end{figure}

\textbf{Slot count adaptivity.} 
The foreground active slots distribution matches the GT object-count distribution (Fig.~\ref{fig:act-distri}), confirming QASA's adaptivity.

\begin{figure}[h]
        \centering
        \includegraphics[width = 0.85\textwidth]{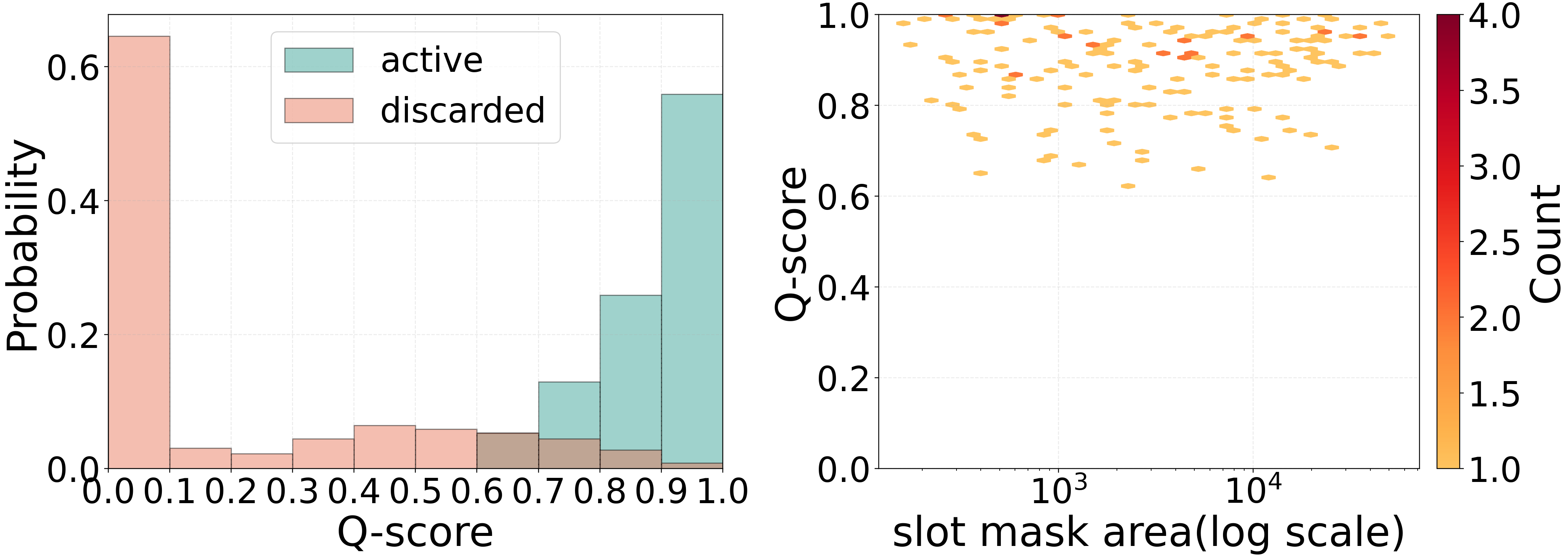}
        \caption{\textbf{Slot-Quality distribution on COCO.}}
        \label{fig:Q-analysis}
    \end{figure}

\textbf{Quality distribution analysis (Fig.~\ref{fig:Q-analysis}).}
The selected and discarded $Q$ distributions are clearly separated, with selected slots concentrated at high scores. Moreover, $Q$ has little correlation with object size, indicating no size bias.

\textbf{Additional time and space cost.} Results in Tab.~\ref{tab:cost} show that our method is efficient.
During training, the additional time and memory overhead is small and remains acceptable.
During evaluation, we do not apply quality-guided gating.
All slots compete for each token, and the extra runtime cost is negligible.

\begin{table}[h]
\centering
\caption{\textbf{Cost comparison.} We report time and per-GPU memory usage for 1-epoch COCO training on 4x RTX 4090s.}
\label{tab:cost}
\begin{tabular}{cccc}
\toprule
Method & Train time & Train space & Eval time \\
\midrule
SPOT & 5m\,27s & 15.8\,GiB & 30\,s \\
Ours & 5m\,47s & 17.2\,GiB & 32\,s \\
\bottomrule
\end{tabular}
\end{table}

\section{Additional Examples}
\label{sec:extra_exam}

\begin{figure*}[thb!]
	\centering
	\includegraphics[width = 0.99 \textwidth]{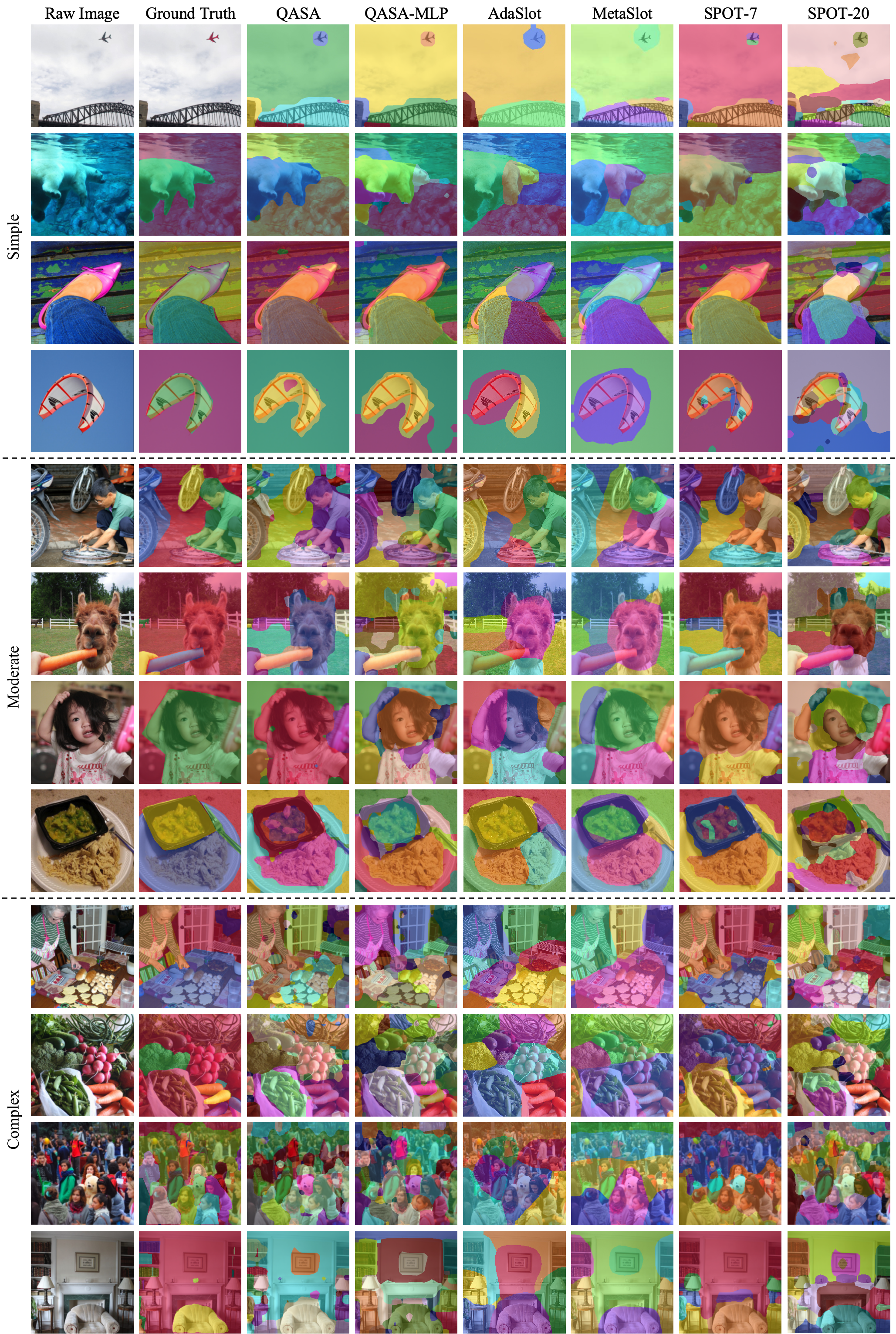}{\\
	\caption{\textbf{Additional examples of \ourmodel~on COCO.}}
	\label{fig:extra-coco}
    }
\end{figure*}

\begin{figure*}[hb!]
	\centering
	\includegraphics[width = 0.95 \textwidth]{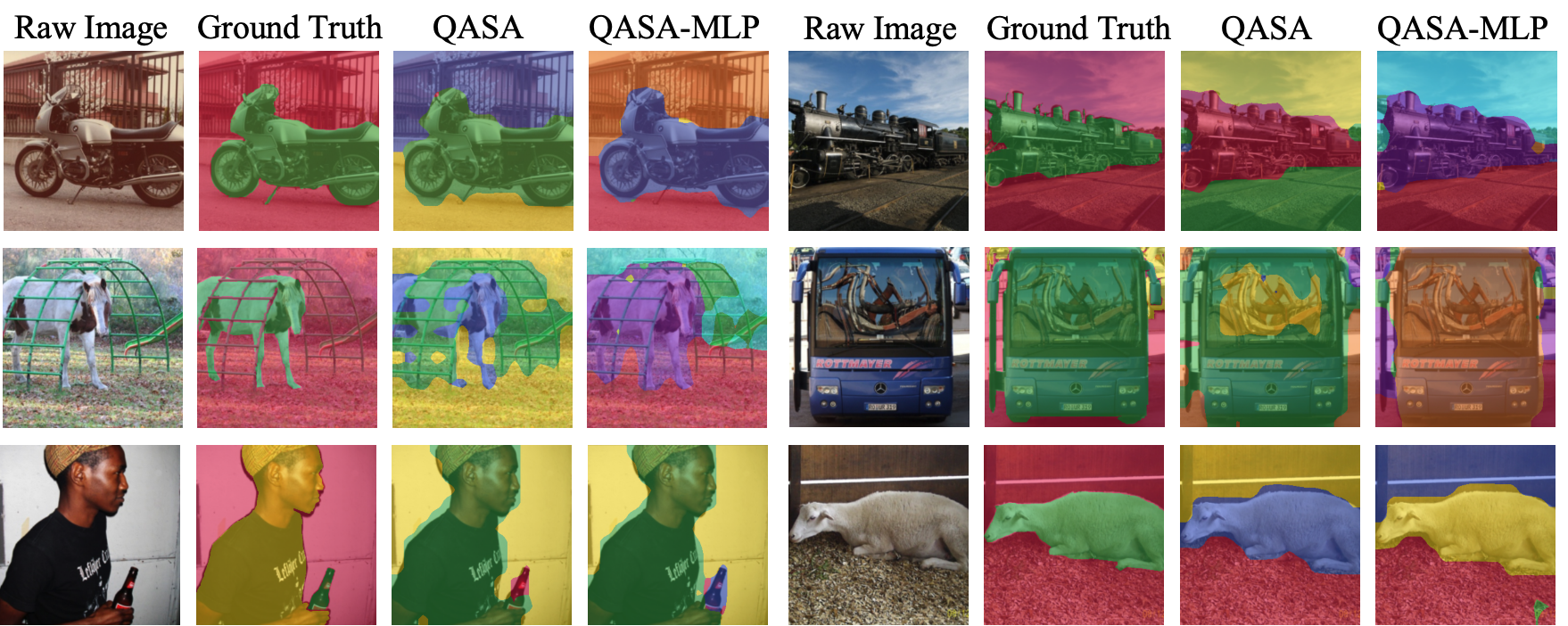}{\\
	\caption{\textbf{Additional examples of \ourmodel~on PASCAL VOC.}}
	\label{fig:extra-voc}
    }
\end{figure*}

\begin{figure*}[thb!]
	\centering
	\includegraphics[width = 0.95 \textwidth]{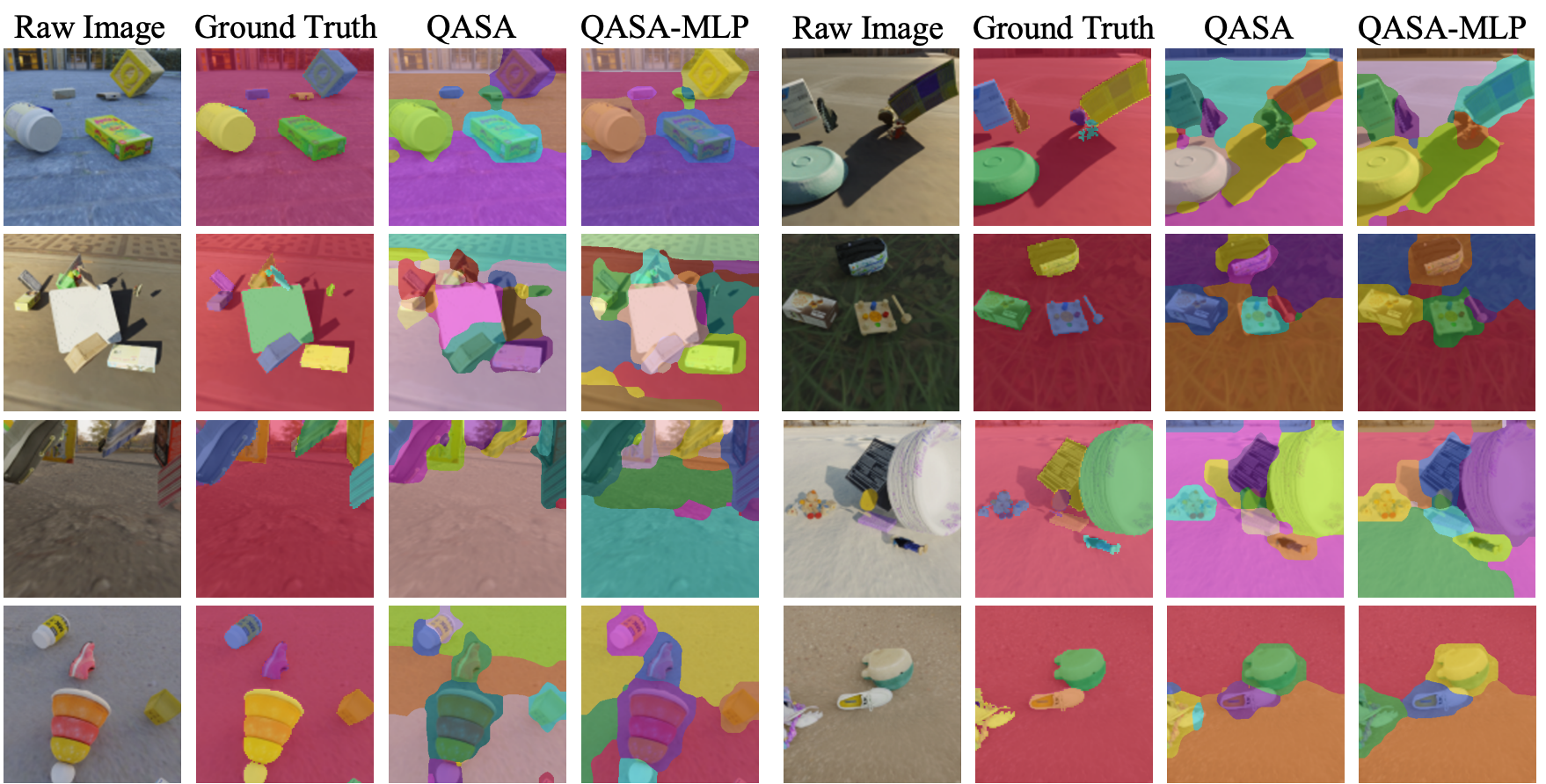}{\\
	\caption{\textbf{Additional examples of \ourmodel~on MOVi-C.}}
	\label{fig:extra-movic}
    }
\end{figure*}

\begin{figure*}[hb!]
	\centering
	\includegraphics[width = 0.95 \textwidth]{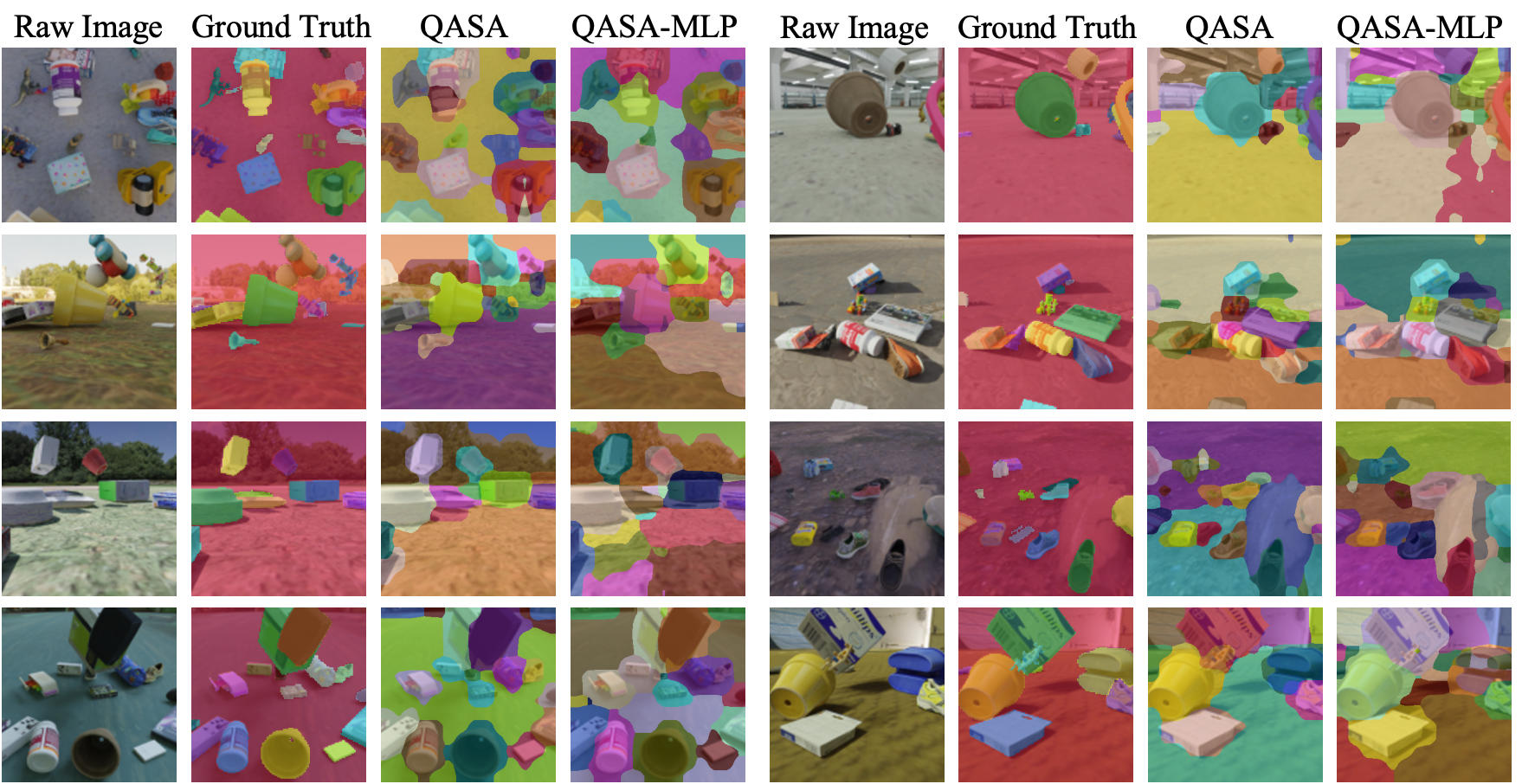}{\\
	\caption{\textbf{Additional examples of \ourmodel~on MOVi-E.}}
	\label{fig:extra-movie}
    }
\end{figure*}

We provide additional visualizations of \ourmodel~on COCO, PASCAL~VOC, and MOVi-C/E, as shown in Figs.~\ref{fig:extra-coco}, \ref{fig:extra-voc}, \ref{fig:extra-movic}, and \ref{fig:extra-movie}.

\end{document}